%% file: root.tex
\documentclass[letterpaper, 10 pt, conference]{ieeeconf}  

\IEEEoverridecommandlockouts                              

\usepackage{graphics} 
\usepackage{epsfig} 
\usepackage{times} 
\usepackage{amsmath} 
\usepackage{amssymb}  

\usepackage{booktabs, multirow} 
\usepackage{float}

\usepackage{algorithm}
\usepackage{algpseudocode}
\usepackage[table]{xcolor}

\usepackage{caption}
\usepackage{subcaption}
\usepackage{pifont}
\newcommand{\cmark}{\ding{51}}%
\newcommand{\xmark}{\ding{55}}%

\usepackage{xcolor}
\usepackage{tablefootnote}

\title{\LARGE \bf

FalconApp: Rapid iPhone Deployment of End-to-End Perception via Automatically Labeled Synthetic Data
}

\author{Yan Miao$^*$, Will Shen$^*$ and Sayan Mitra 
\thanks{$^*$These authors contributed equally to this work}
\thanks{{\footnotesize The authors are with the Department of Electrical and Computer Engineering, University of Illinois at Urbana Champaign {\tt\footnotesize \{yanmiao2, wshen15, mitras\}@illinois.edu}}}
}

\begin{document}
\bstctlcite{BSTcontrol}

\raggedbottom

\IEEEaftertitletext{%
\begin{center}
\begin{minipage}{\textwidth}
\centering
\includegraphics[width=0.95\textwidth]{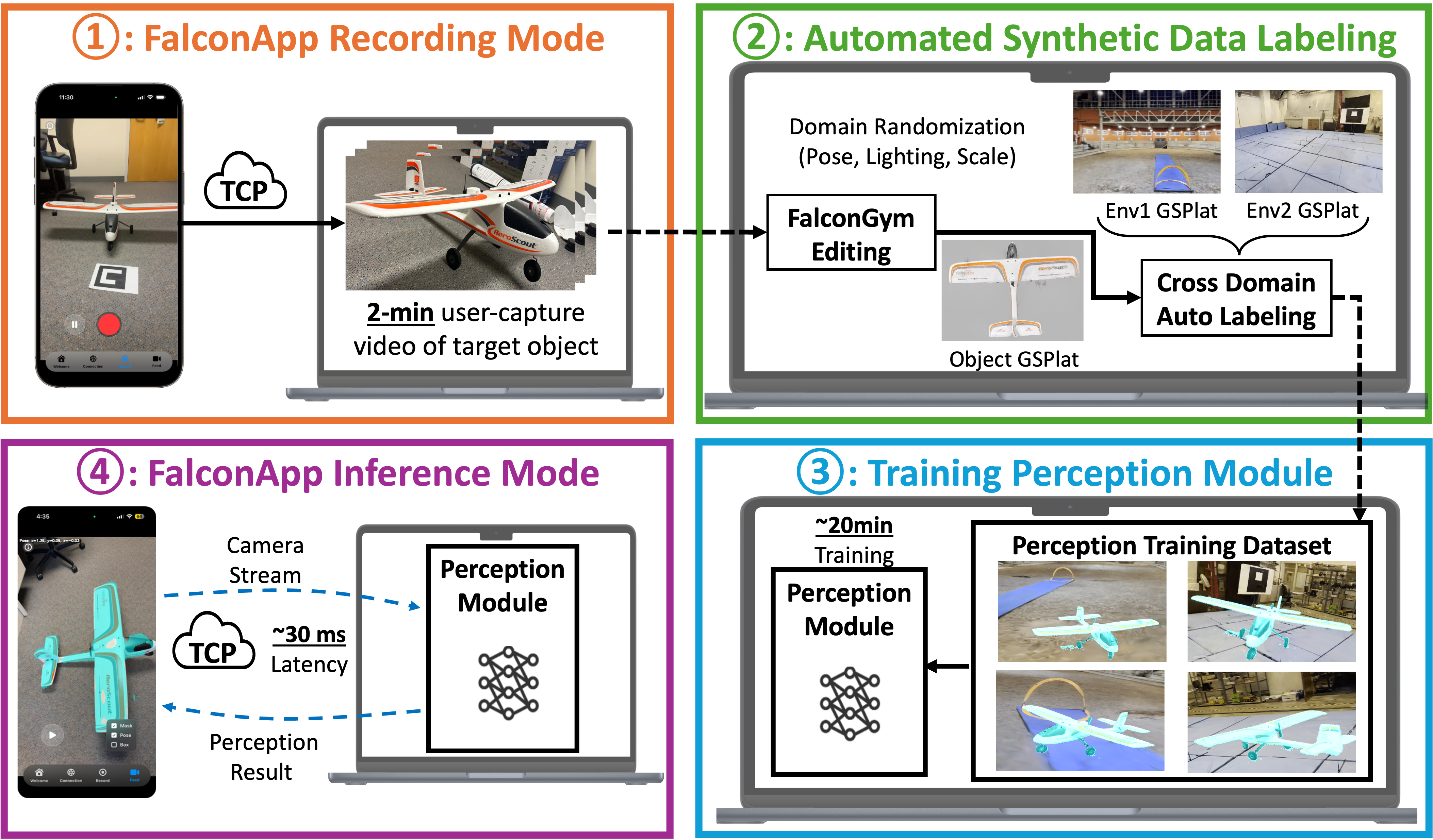}
\captionof{figure}{\small{
\textbf{FalconApp Pipeline:}
{\small }
From a 2-minute iPhone video of an object, FalconApp reconstructs a photorealistic GSplat, composites it with diverse background in FalconGym 2.0 \cite{MiaoEtAl:ICRA26}, automatically synthesizes ground-truth masks and pose under randomized viewpoints, produces a full perception module in $\sim$20 min and returns it to the app frontend for live inference with an overall latency of $\sim$30\,ms.
}}
\label{fig:FalconApp-pipeline}
\end{minipage}
\end{center}
}

\maketitle

\thispagestyle{empty}
\pagestyle{empty}


\begin{abstract}
Reliable perception for robotics depends on large-scale labeled data, yet real-world datasets rely on heavy manual annotation and are time-consuming to produce.
We present FalconApp, an iPhone app with an end-to-end frontend-backend pipeline that turns a short handheld capture of a rigid object into a perception module for mask detection and 6-DoF pose estimation.
Our core contribution is a rapid mobile deployment pipeline paired with a photorealistic auto-labeling workflow: from a user-captured video of an object, FalconApp reconstructs an editable GSplat asset, composites it with diverse photorealistic backgrounds, renders synthetic images with ground-truth masks and poses, trains the perception module, and deploys it back to the iPhone frontend.
Experiments across five rigid objects with diverse geometry and appearance show that FalconApp produces usable perception models with about 20 minutes of synthetic-data generation and training per object on average, around 30\,ms end-to-end on-device latency on iPhone, and better overall pose accuracy than a PnP baseline on 4 / 5 objects in both simulation and real-world evaluation.
\end{abstract}

\input{1-introduction}
\input{2-related_work}
\input{3-method}

\input{4-experiment}
\input{5-discussion}


\bibliographystyle{IEEEtran}
\bibliography{reference}










\end{document}

%% file: 1-introduction.tex
\section{Introduction}

\begin{figure*}[htbp]
    \centering
    \includegraphics[width=0.9\linewidth]{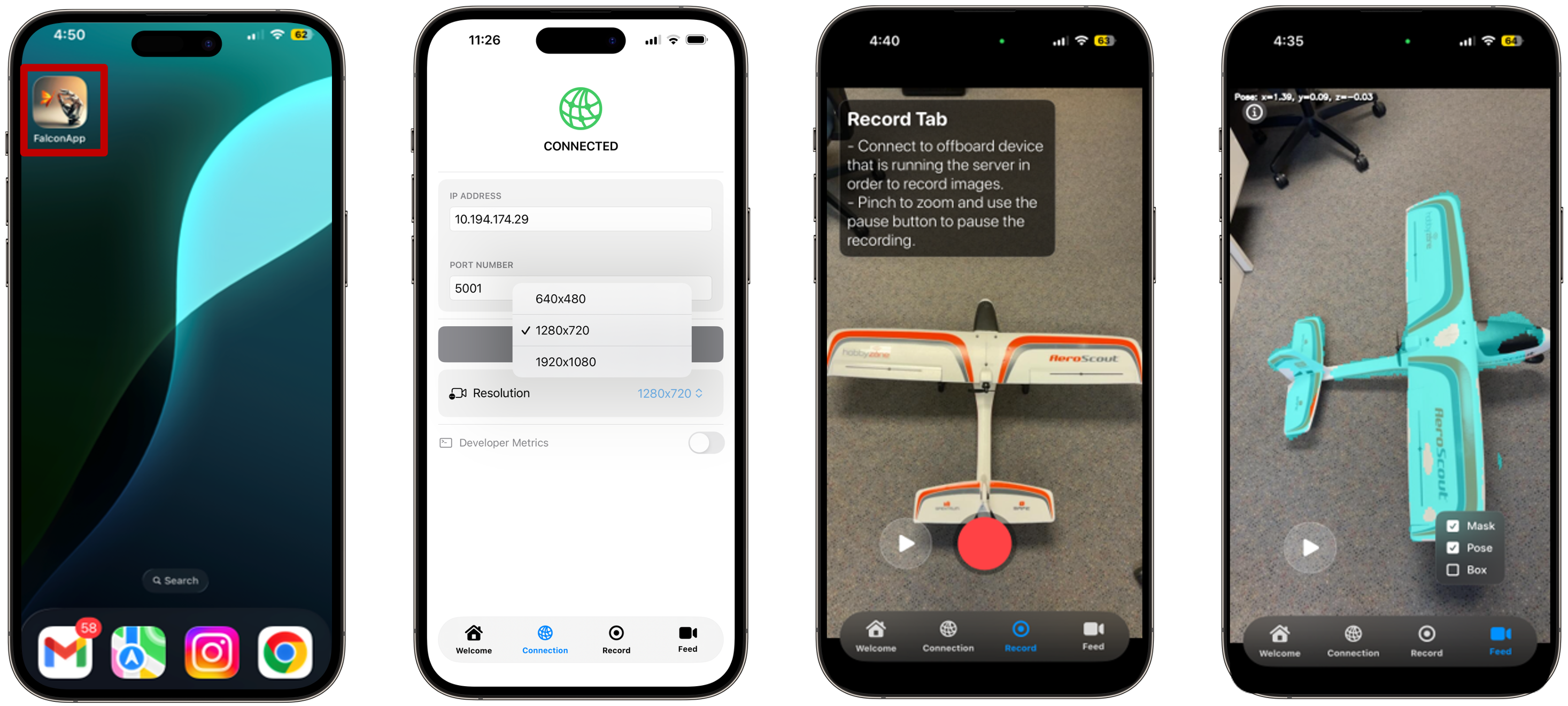}
    \caption{\small{\textbf{FalconApp frontend interface:} the four views, from left to right, show iPhone app installation, TCP connection to the backend, recording mode (Stage~1), and inference mode (Stage~4) in the pipeline of Figure~\ref{fig:FalconApp-pipeline}.}}
    \label{fig:app-interface}
\end{figure*}

Reliable perception modules are central to many robotics tasks, including manipulation, navigation, and autonomous driving.
However, training a reliable perception module (such as Mask R-CNN and PoseCNN \cite{8237584,xiang2018posecnn}) typically requires large real-world labeled datasets (such as ImageNet and BDD100K \cite{5206848,bdd100k}). 
Yet collecting those labels is often slow and expensive because it relies heavily on manual annotation.
Synthetic data from simulation offers an attractive alternative because labels can be generated automatically, but traditional simulators such as CARLA \cite{Dosovitskiy17} and Gazebo \cite{1389727} often lack photorealism and can therefore introduce a large sim-to-real gap on real robots.
Recent advances in NeRF \cite{10.1145/3503250} and 3D Gaussian Splatting (GSplat \cite{kerbl3Dgaussians}) suggest a more photorealistic route to automatic labeling, but they do not by themselves provide an end-to-end workflow from casual user videos to usable perception modules on real devices.

We address this gap with FalconApp, an iPhone app with a frontend-backend pipeline for rapidly developing a perception module for mask detection and 6-DoF pose estimation.
As illustrated by Figure~\ref{fig:FalconApp-pipeline}, FalconApp's main novelty is to couple a rapid mobile deployment pipeline with a photorealistic auto-labeling workflow: it records a short handheld iPhone capture, reconstructs an editable object GSplat, generates cross-domain synthetic RGB images with aligned masks and poses, trains a perception module, and returns it to the iPhone frontend for live inference on unseen real-world images.
The measured synthetic-data generation and training stages finish in about 20 minutes on average and support practical object-specific Sim-to-Real deployment.
We evaluate FalconApp on five rigid objects with diverse appearance, scale, and symmetry, and show about 30\,ms end-to-end inference latency together with stronger overall pose accuracy than a PnP baseline.

In summary, our key contributions are: (1) a rapid mobile deployment pipeline that turns a short iPhone capture into an on-device object-specific perception module; and (2) a photorealistic auto-labeling workflow based on editable GSplats that generates synthetic RGB images with aligned mask and pose labels, enabling training without manual annotation.

%% file: 3-method.tex
\section{Methodology}

\begin{figure*}[htbp]
    \centering
    \includegraphics[width=\linewidth]{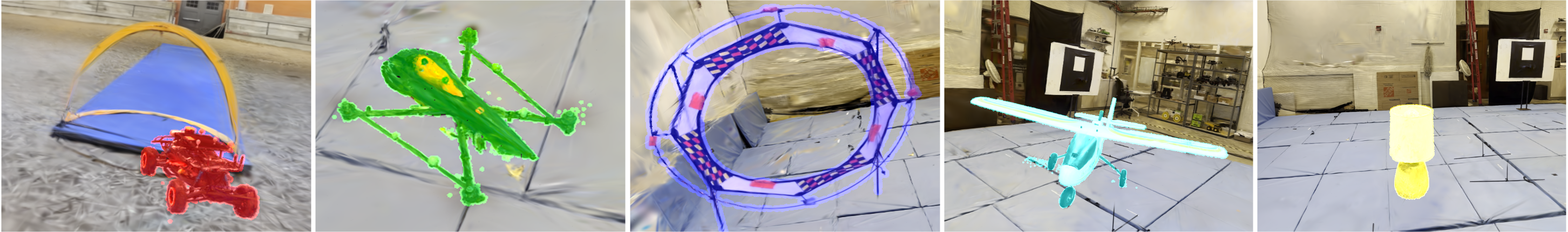}
    \caption{\small{\textbf{Demonstration of Automatically labeled Synthetic Data:} FalconApp produces photorealistic synthetic images and automatically generates labels for objects with diverse geometry and appearance to train a perception module.}}
    \label{fig:five-labeled-objects}
\end{figure*}

FalconApp couples a rapid mobile deployment pipeline with a photorealistic auto-labeling workflow, turning a short iPhone capture of a rigid object into a trained perception module for mask detection and 6-DoF pose estimation, with synthetic-data generation and training taking about 20 minutes per object on average.
As summarized in Figure~\ref{fig:FalconApp-pipeline}, the pipeline has four stages.
In Section~\ref{sec:stage-one-frontend}, the mobile frontend records a short handheld video and uploads the capture to the backend.
In Section~\ref{sec:stage-two-auto-label}, the backend reconstructs a photorealistic GSplat asset, edits it in FalconGym 2.0, and renders cross-domain synthetic images with automatic mask and pose labels.
In Section~\ref{sec:stage-three-perception-training}, the backend trains a perception module from the generated data and packages the resulting model.
Finally, in Section~\ref{sec:stage-four-inference}, the deployed model is returned to the phone for live image-based inference.

\subsection{FalconApp Frontend: Recording Mode}
\label{sec:stage-one-frontend}

\begin{figure}[htbp]
    \centering
    \includegraphics[width=\linewidth]{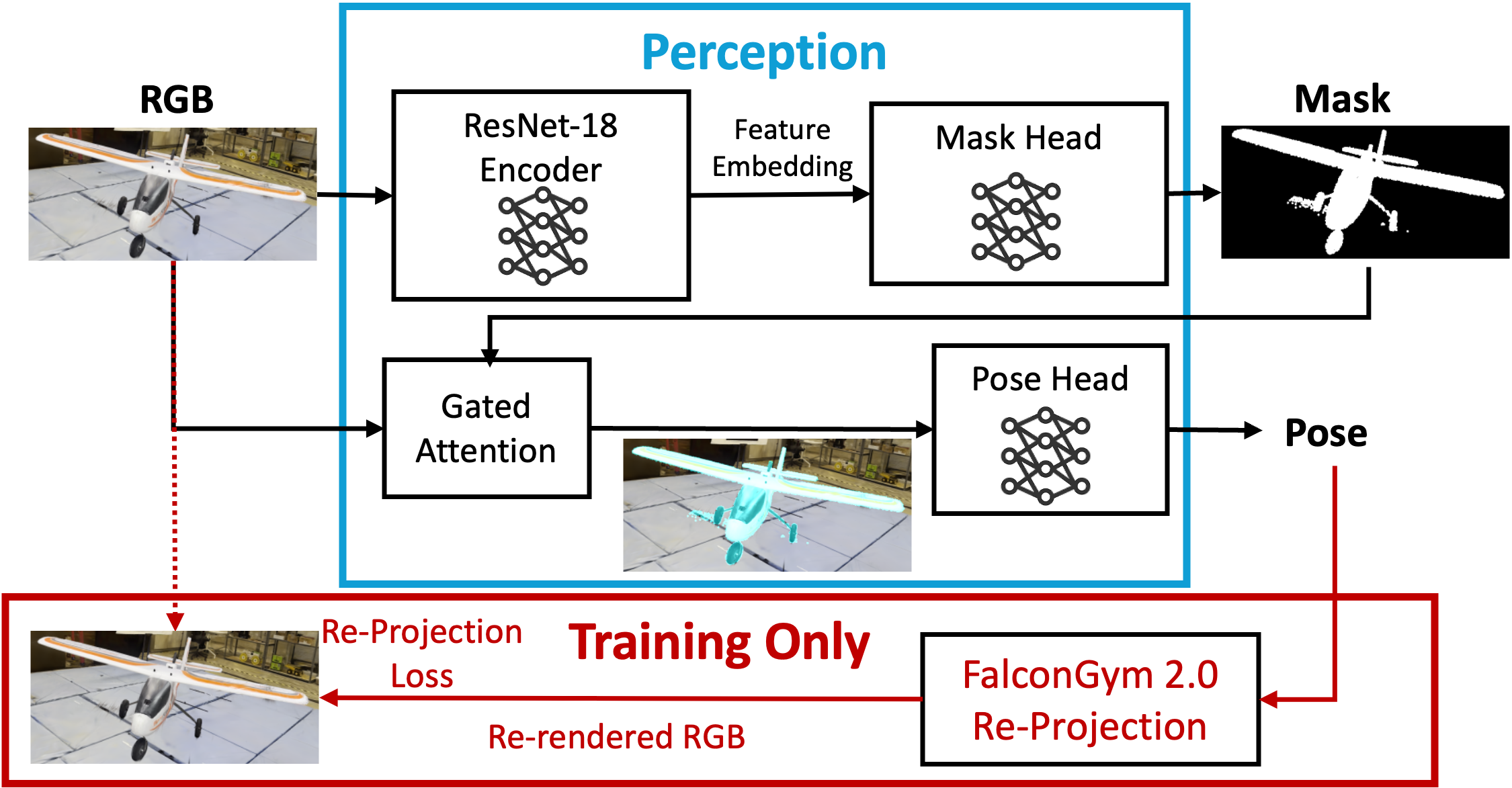}
    \caption{\small{
    \textbf{Perception Architecture:}
    an encoder extracts features that feed into a mask prediction head;
    the predicted mask conditions a gated-attention pose head to regress the relative 6-DoF target pose.
    During training, we use FalconGym 2.0 to re-render targets at predicted poses and apply reprojection loss to improve geometric consistency and pose estimation.
    }}
    \label{fig:pose-estimator}
\end{figure}

FalconApp is implemented in Swift and deployed directly on an iPhone, with the user interface shown in Figure~\ref{fig:app-interface}.
The frontend provides a lightweight capture interface for recording a short RGB video while the user walks around the object to obtain multi-view coverage.
Captured video and associated camera metadata are sent to the backend through a TCP client, enabling direct frontend-to-backend upload without manual file transfer.
The same mobile interface also displays a welcome page with instructions and the live inference outputs described in Section~\ref{sec:stage-four-inference}.

\subsection{FalconApp Backend: Automated Synthetic Data Labeling}
\label{sec:stage-two-auto-label}

After upload, the backend reconstructs a 3D Gaussian Splatting (GSplat) \cite{kerbl3Dgaussians} representation of the target from the handheld capture using FalconGym \cite{11247178}.
We then use FalconGym 2.0's Edit API \cite{MiaoEtAl:ICRA26} to separate the object splats from the capture background and obtain an editable object GSplat.
To generate training data, the object GSplat is composited with multiple background GSplats and rendered under randomized viewpoints, object scales, and lighting conditions.
These domain randomization techniques are used to minimize sim-to-real gaps when deploying on iPhone with unseen images.
Because FalconGym 2.0 exposes the object transform and camera intrinsics/extrinsics, each synthetic image is automatically paired with a pixel-accurate foreground mask and the target's relative 6-DoF pose in the camera frame.
We also render empty-scene images to improve robustness when the target is partially or fully out of view and to reduce false positives.
Examples of the auto-generated masks are shown in Figure~\ref{fig:five-labeled-objects}. On average, synthetic rendering and auto-label process take about 0.2s per $1280\times720\times3$ RGB image on an RTX~4090 GPU.

\subsection{FalconApp Backend: Perception Module Training}
\label{sec:stage-three-perception-training}

Each uploaded object receives its own perception module. Readers are referred to FalconTrack \cite{miao2026falcontrackphotorealisticautolabeledperception} for details.
As shown in Figure \ref{fig:pose-estimator}, the module first predicts a foreground mask, then applies gated attention over the mask and RGB features before feeding them into a pose head to suppress background clutter while preserving appearance cues needed for orientation.
Training follows a short staged schedule: we first optimize mask prediction, and then jointly optimize mask and pose using the synthetic RGB, mask, and pose labels.
The final objective combines a segmentation loss, pose regression loss, and a reprojection consistency term that re-renders the object in FalconGym 2.0 at the predicted pose and compares it with the input image using SSIM.
Pose and reprojection losses are applied only when object is in-view.

\subsection{FalconApp Frontend: Deploy Perception on iPhone}
\label{sec:stage-four-inference}
Once training completes, the backend packages the target-specific model and returns it to the mobile client.
The frontend then loads the trained estimator and runs it on incoming camera images for live segmentation and 6-DoF pose prediction.
For each frame, the app resizes the RGB image to the training resolution, performs a forward pass, and visualizes the predicted mask together with the relative pose estimate, as shown in the rightmost subfigure of Figure~\ref{fig:app-interface}.
This closes the end-to-end loop: a short casual phone capture is converted into a perception module that can be used immediately for live object-specific inference.

%% file: 4-experiment.tex
\section{Experiments}

We evaluate FalconApp on five rigid objects that span different geometry, scale, and symmetry: a car, a quadrotor, a gate, a plane, and a lamp.
We report both pipeline deployment efficiency and perception accuracy.

\subsection{Pipeline Runtime and Reconstruction Quality}
\label{sec:latency-analysis}
We first evaluate the object GSplats reconstructed in Section~\ref{sec:stage-two-auto-label} together with FalconApp's measured backend runtime and on-device inference latency.
For reconstruction quality, we report the standard photorealism metrics SSIM, PSNR, and LPIPS.
We then report the number of synthetic images, automatic labeling time, training time, and end-to-end inference latency for each object.

\begin{table*}[htbp]
\centering
\caption{\small{Reconstruction quality, backend runtime, and on-device inference latency across five objects. 
}}
\label{tab:FalconApp-pipeline-metrics}
\resizebox{\textwidth}{!}{
\scriptsize
\begin{tabular}{lcccc|cccc}\toprule
    & \multicolumn{4}{c|}{Object GSplat Photorealism} & \multicolumn{4}{c}{Deployment Metrics} \\\midrule
    Object & \# Gaussians & SSIM$\uparrow$ & PSNR$\uparrow$ & LPIPS$\downarrow$ & \# Synthetic Images & Auto-Label Time & Training Time & Total Latency \\\midrule
    Car & 20 k & 0.88 & 31.8 & 0.09 & 1000 & 2 min & 10 min & 30 ms \\
    Quadrotor & 23 k & 0.85 & 32.4 & 0.06 & 1000 & 2 min & 10 min & 32 ms \\
    Gate & 74 k & 0.71 & 30.9 & 0.12 & 1000 & 8 min & 15 min & 30 ms \\
    Plane & 38 k & 0.73 & 30.0 & 0.08 & 1000 & 5 min & 15 min & 31 ms \\
    Lamp & 16 k & 0.77 & 30.7 & 0.08 & 1000 & 2 min & 10 min & 29 ms \\
\bottomrule
\end{tabular}
}
\end{table*}

As shown in Table~\ref{tab:FalconApp-pipeline-metrics}, the reconstructed GSplats are sufficiently photorealistic to support downstream synthetic data generation across all five objects.
Generating 1000 synthetic images and training a perception module takes under 20 minutes per object on average, while end-to-end total latency including inference and TCP communication remains around 30\,ms.
These results suggest that FalconApp is fast enough to run perception with a moving camera at around 30 fps. 

\subsection{Perception Module Accuracy}
\label{sec:accuracy-analysis}
We next evaluate FalconApp's perception module on the same five objects and compare it with a traditional perspective-$n$-point (PnP) baseline \cite{7368948}.
PnP is a reasonable conventional baseline here because it is a standard instance-specific 6-DoF pose estimator for known rigid objects, but unlike FalconApp it still relies on manually selected correspondences rather than an end-to-end learned auto-labeling pipeline.
We report mask intersection-over-union (IoU), mean translational error (MTE), and mean angular error (MAE).
To compare across objects with different sizes, MTE is normalized by object size and reported as a percentage, while MAE is reported in radians.
Because PnP outputs only pose estimates and still requires manual feature selection, we leave IoU blank for that baseline.

We first evaluate both methods on 200 unseen FalconGym 2.0 images per object across two environments.
We then repeat the evaluation on 200 real images with different background and use an ArUco marker-based calibration procedure to recover the ground truth pose in the camera frame.
Because ground truth real-world masks are unavailable without manual annotation, real-world IoU is omitted.

\begin{table*}[htbp]
\centering
\caption{\small{Mask and pose estimation accuracy on five objects in FalconGym 2.0 and the real world. 
}}
\label{tab:FalconApp-perception-result}
\scriptsize
\resizebox{0.8\textwidth}{!}{
\begin{tabular}{lcc|ccc|ccc}\toprule
    & & & \multicolumn{3}{|c|}{\textbf{FalconGym 2.0 (2 Environments)}} & \multicolumn{3}{c}{\textbf{Real World (2 Environments)}} \\\midrule
    Object & Method & Auto-Labels? & IoU$\uparrow$ & MTE (\%)$\downarrow$ & MAE (rad)$\downarrow$ & IoU$\uparrow$ & MTE (\%)$\downarrow$ & MAE (rad)$\downarrow$ \\\midrule
    Car & PnP \cite{7368948} & \xmark & - & 52\% & 0.88 & - & 82\% & 1.21 \\
    & FalconApp & \cmark & 0.83 & 24\% & 0.27 & - & 43\% & 0.83 \\
    \midrule
    Quadrotor & PnP \cite{7368948} & \xmark & - & 71\% & 1.12 & - & 82\% & 0.82 \\
    & FalconApp & \cmark & 0.79 & 32\% & 0.31 & - & 41\% & 0.37 \\
    \midrule
    Gate & PnP \cite{7368948} & \xmark & - & 60\% & 0.76 & - & 72\% & 0.69 \\
    & FalconApp & \cmark & 0.72 & 41\% & 0.41 & - & 52\% & 0.52 \\
    \midrule
    Plane & PnP \cite{7368948} & \xmark & - & 37\% & 0.62 & - & 57\% & 0.66 \\
    & FalconApp & \cmark & 0.81 & 24\% & 0.32 & - & 34\% & 0.44 \\
    \midrule
    Lamp & PnP \cite{7368948} & \xmark & - & 62\% & 0.98 & - & 72\% & 1.1 \\
    & FalconApp & \cmark & 0.62 & 71\% & 0.84 & - & 85\% & 1.4 \\
\bottomrule
\end{tabular}
}
\end{table*}

Table~\ref{tab:FalconApp-perception-result} shows that FalconApp achieves strong mask and pose performance on most objects and improves pose accuracy over PnP on four of the five targets in both simulation and the real world.
The lamp remains the hardest case because its rotational symmetry makes orientation ambiguous, which increases angular error and also weakens translation estimates.
Overall, the results suggest that FalconApp's synthetic-data pipeline transfers well enough for practical target-specific deployment.

%% file: 5-discussion.tex
\section{Conclusion \& Discussion}

We presented FalconApp, an iPhone app with an end-to-end frontend-backend pipeline that turns a short handheld capture of a rigid object into a perception module for mask detection and 6-DoF pose estimation.
%
%
Experiments on five objects show that FalconApp produces usable perception modules within about 20 minutes, runs with around 30\,ms end-to-end latency on iPhone, and achieves better overall pose accuracy than a PnP baseline on four of the five objects.

The current pipeline still assumes rigid known objects, depends on synthetic-to-real transfer quality, and remains weakest on highly symmetric shapes such as the lamp.
Future work includes stronger domain randomization, improved handling of symmetry and ambiguity, fully onboard deployment, and tighter integration with downstream robotics tasks such as aerial navigation \cite{miao2025falconwingultralightindoorfixedwing} and manipulation.

